\documentclass{article}

\usepackage{lastpage}
\usepackage{epsfig}

\def\cdd{\mbox{\boldmath$\cdot$}~}
\makeatletter
\def\@oddfoot{\hfill}

\newcount\shumeicount
\def\setshumei#1#2#3{%
  \shumeicount=\count0
  \def\@oddhead{%
    \raise-5pt\hbox to0pt{\vrule width\hsize height 0pt depth 0.4pt\hss}\relax
    \ifnum \shumeicount=\count0
      \raise-7pt\hbox to0pt{\vrule width\hsize height 0pt depth 0.4pt\hss}\relax
      #1
    \else
      \ifodd\count0
        #2
      \else
        #3
       \fi
     \fi
  }%
}
\makeatother

\makeatletter
\def\@oddfoot{\hfill}
\newcount\shujiaocount
\def\setshujiao{%
  \shujiaocount=\count0
  \def\@oddfoot{%
      \ifodd\count0
      \else
      \fi
  }%
}
\makeatother

\def\biaoti#1#2#3#4{{
  \vspace*{0.3cm}
  \begin{flushleft} \Large\bf #1\end{flushleft}
  \vspace*{-0.2cm}
      \begin{flushleft}
      \bf #2
      \end{flushleft}
      \footnotetext{\hspace{-6mm} #3\\ #4}}}

\def\dshm#1#2#3#4
{\setshumei{ Jrl Syst Sci \& Complexity (#1) #2:
{\thepage--\pageref{LastPage}}\hfill}
            {\hfill {\small #3}\hfill\hbox to0pt{\hss\thepage}}
            {\hbox to0pt{\thepage\hss}\hfill {\small #4}\hfill
            }
            \setshujiao}
\def\drd#1
{{\vskip 1cm\small \begin{flushleft}
 #1 \\
\copyright 2006 Springer Science + Business Media, Inc.
\end{flushleft}}}
\def\dab#1#2{\noindent {{\small\bf Abstract~~}}{{\small #1}}
            \vskip 0.1cm
             \noindent{{\small\bf Key words~~}}{{\small #2}}
                      }
\def\dse#1{\vskip 0.6cm\noindent
        {\large\bf #1}
        \vskip 0.4cm}
\def\rf#1#2{\parindent=0pt\hangindent=0.6cm\hangafter=1\small
            \parbox[t]{0.6cm}{[#1]}#2\par}
\def\rfne{\vskip 0.5cm  \centerline{\bf References} \vskip 0.5cm
               \parindent 0pt}
  
\def\dsee#1{\vskip 0.3cm \noindent{\bf #1} \vskip 0.2cm}

\input{amssym.def}
\usepackage{graphicx}

\begin{document}


\biaoti{PREDICTING RELEVANT EMPTY SPOTS IN SOCIAL INTERACTION}{Yoshiharu MAENO \cdd Yukio OHSAWA}{Yoshiharu MAENO \\ {\it Graduate School of Systems Management, Tsukuba University, Tokyo} $1120012$, {\it Japan}. \\ Email: maeno.yoshiharu@nifty.com. \\ Yukio OHSAWA \\ {\it School of Engineering, University of Tokyo, Tokyo} $1138656$, {\it Japan}. \\ Email: ohsawa@q.t.u-tokyo.ac.jp.}{}

\drd{Received: August 15, 2007}


\dshm{*}{*}{EMPTY SPOTS IN SOCIAL INTERACTION}{Yoshiharu MAENO $\cdot$ Yukio OHSAWA}

\dab{An empty spot refers to an empty hard-to-fill space which can be found in the records of the social interaction, and is the clue to the persons in the underlying social network who do not appear in the records. This contribution addresses a problem to predict relevant empty spots in social interaction. Homogeneous and inhomogeneous networks are studied as a model underlying the social interaction. A heuristic predictor function method is presented as a new method to address the problem. Simulation experiment is demonstrated over a homogeneous network. A test data set in the form of market baskets is generated from the simulated communication. Precision to predict the empty spots is calculated to demonstrate the performance of the presented method.}{Communication, Empty spot, Predictor function, Social interaction, Social network}


\dse{1~~Problem - empty spot}

The activity of an organization is often under the influence of hidden but relevant persons. The activity refers to decision-making and action-taking. In this contribution, the word {\it hidden} means unobserved (unobservable) or invisible by the method used in the observation, rather than anonymous. The invisibility of the hidden persons is caused from the limited capability of the observation method, or from the limited prior understanding of the targets to observe. The organization may be a family, school, company, community, society etc.

For example, a financially supporting conspirator, who provides commanders with money, communication means, or weapons, is often hidden behind terrorism attacks. Commanders would appear one after another if such a conspirator were not detected and arrested. It is, therefore, critical to predict the presence of such hidden but relevant persons from the observed records on the social interactions of an organization. Based on the prediction, we can invent a scenario for proactive investigation. The scenario turns the threat of disasters from the terrorism attacks into the opportunity to discover and destroy the hidden conspirator's social network supporting terrorism.

We define the above problem more specifically with three ideas. They are {\it social interaction}, {\it social network}, and {\it empty spot}.

\begin{itemize}

\item{} A social interaction is a dynamic influence dissemination among persons through conversation, meeting, collective action etc. Communication is of particular interest since it often takes place on the electronic media, which is one of the major targets of surveillance. 

\item{} An organization can be modeled as a social network which underlies below the social interaction. Nodes are persons. Links are relationship such as friendship, business partnership, chain of command etc. The links can be undirectional, unidirectional, or bidirectional. Variety of network topologies are known. A scale-free network$^{[3]}$ and a small-world network$^{[19]}$ were studied mathematically in detail. The topology of the real networks are diverse. The topologies of contemporary inter-working terrorists, self-organizing on-line community, and purposefully organized business team do not resemble.

\item{} The empty spot in the social interaction is the main topic of this contribution. It refers to an empty hard-to-fill space, which can exist in the observed records of the social interaction, and is the potential clue to the persons in the underlying social network who do not appear in the records. Such hidden persons are the origin of the empty spot in a nutshell.

\end{itemize}

In this contribution, the problem we address is to discover relevant empty spots in a complex social interaction. We propose a {\it heuristic predictor function method} to predict the relevant empty spots and the hidden persons from communication records. The method is presented in detail in 4 after studying the related works in 2 and the network models (homogeneous and inhomogeneous network) in 3. Simulation experiment is demonstrated in 5. A test data set is generated in the form of market baskets as the simulated communication records over a homogeneous network. Precision to discover the empty spots is calculated to evaluate the performance of the method for three trial cases. Concluding remarks are presented in 6.

\dse{2~~Related Work}

The problem is related to a node discovery problem$^{[14],[16]}$. Expertise in computer and social sciences is significant. Complex networks such as scale-free networks (Barab\'{a}si-Albert model$^{[3]}$) and small-world networks (Watts-Strogatz model$^{[19]}$) presented us insight on the structure and evolution of a large-scale network. Scientists' collaboration or actors in movies are examples. Power law governing the scale-free network appears from the preferential attachment of the nodes in a growing network. A few very big hub nodes emerge, accounting for the winner-takes-all phenomena. The small-world network is highly clustered like a regular lattice, but has small diameter like a random graph. The name originates in the small-world phenomena known as six degree of separation.

Search efficiency in a network$^{[1]}$ is of particular interest for practical applications. The hub nodes in scale-free networks are useful in designing local search strategies. The hub nodes improve the efficiency to access relevant nodes. Error and attack tolerance were studied$^{[2]}$. Scale-free networks display a surprisingly high degree of tolerance against random errors. Other networks do not have such a property. The error tolerance is, however, at the expense of attack survivability. The network is broken into many isolated fragments when the hub nodes are targeted. Centrality and brokerage measures$^{[4], [5]}$ are convenient values summarizing the network topology. Degree, betweenness, closeness, and eigenvector centrality are popular among them. They have been studied in many cases of the social network analysis.

Empirical study on social networks uncovered many aspects of the criminal and terrorist organizations in the past and at present$^{[17]}$. Intelligence sharing, knowledge management, and simulation techniques are described as well as the social network analysis. Hidden Markov model and Bayesian network are applied to predicting terrorist attacks$^{[18]}$. Activities are modeled and patterns of anomalous behavior are identified. It helps intelligence analysts connect the fragmented facts more quickly. P. S. Keila {\it et al.} applied factor analysis (singular value decomposition and semi-discrete decomposition) to study email exchange in an American energy company, Enron Corporation, which ended in bankruptcy due to the institutionalized accounting fraud in 2001$^{[7]}$. The word use in the emails is correlated to the function within the organization. The word use among those involved in the alleged criminal activity is distinctive.

Criminal organizations tend to be strings of inter-linked small groups that lack a central leader but coordinate their activities along logistic trails and through bonds of friends, and that hypothesis can be built by paying attention to remarkable white spots and hard-to-fill positions in a network$^{[8]}$. The white spots and hard-to-fill positions in the observed social interactions correspond to the empty spots in this contribution. V. E. Krebs investigated the 9/11 terrorist network in 2001$^{[9]}$ . It reveals the relevance of conspirators who reduce the distance between hijackers, enhance communication efficiently, and act as a conduit for money and knowledge. The 9/11 terrorist network was also investigated$^{[15]}$ from the viewpoint of efficiency and security trade-off by analyzing the change in degree, betweenness and closeness centrality measures. It is suggested that more security-oriented structure arises from longer time-to-task of the terrorists' objectives. Conspirators improve communication efficiency, preserving hijackers' small visibility and exposure.

\dse{3~~Network Model}

Social network modeling$^{[4]}$ is a basic tool to describe an organization like a family, school, company, community, society etc. A node represents a person. A link represents relationship between two persons, which we simply assume is equal to the presence of communication between the persons. Information dissemination with the communication is a social interaction. An origin of complexity of the social interaction is the complexity and variability of the underlying network topologies. Different networks describe different organizations. We can classify the networks into two classes. They are inhomogeneous networks and homogeneous networks.

The definition is as follows. A homogeneous network refers to one that consists of the nodes which have similar local topological characteristics. The network is homogeneous when the variance of the nodal degree ($\sigma(d)/\mu(d)$) is small. The standard deviation and mean of the nodal degree $d$ are denoted by $\sigma(d)$ and $\mu(d)$. Many random networks including an Erd\"{o}s-R\'{e}nyi model and a Watts-Strogatz model are homogeneous. On the other hand, an inhomogeneous network refers to one that consists of the nodes which have variable local topological characteristics ($\sigma(d)/\mu(d)$ is large). A strong center-and-periphery structure or a leader-follower relationship is the indicator of an inhomogeneous network.

\dsee{3.1~~Inhomogeneous network}

\begin{figure}
\begin{center}
\includegraphics[scale=0.5,angle=-90]{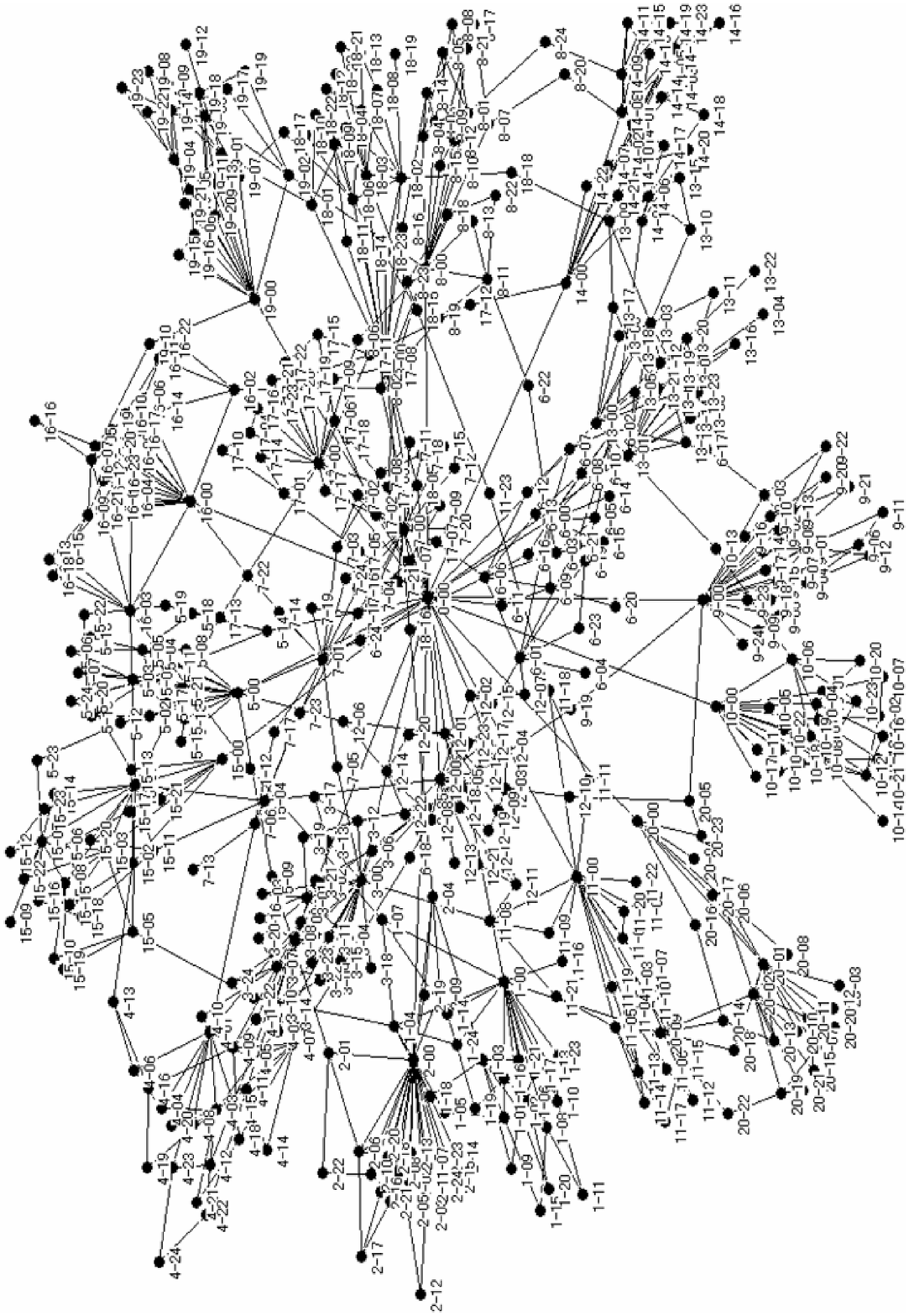}
\end{center}
\caption{Example of an inhomogeneous network consisting of $|n|=490$ nodes. It is a scale-free network governed by a power low (Barab\'{a}si-Albert model). Center-and-periphery structure is evident.}
\label{FIGURE1}
\end{figure}

\begin{figure}
\begin{center}
\includegraphics[scale=0.3,angle=-90]{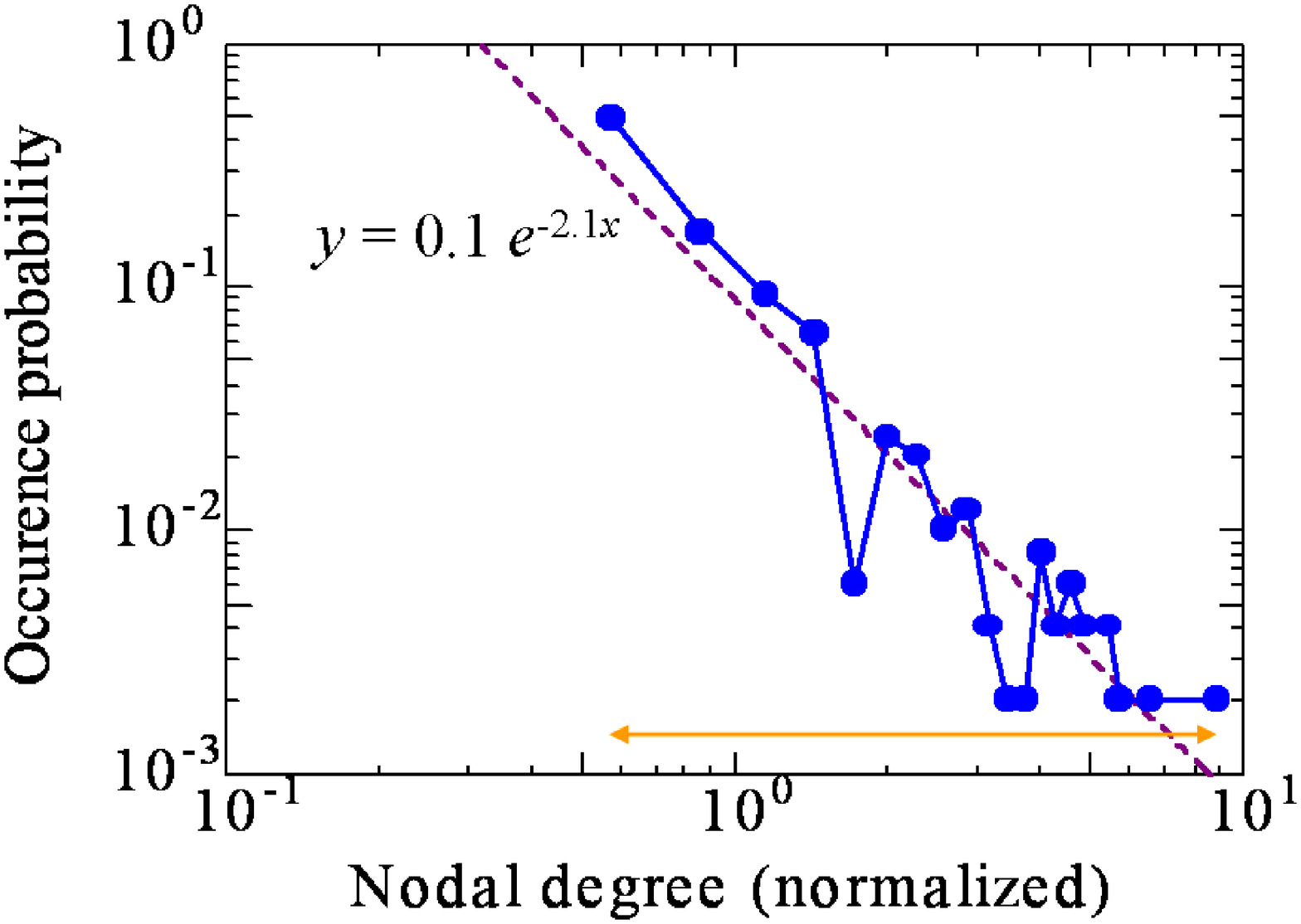}
\end{center}
\caption{Occurrence probability distribution $P(d)$ of the nodal degree $d$ as a function of the normalized degree $d/\mu(d)$.}
\label{FIGURE2}
\end{figure}

A typical inhomogeneous network is a scale-free network derived from a Barab\'{a}si-Albert model$^{[3]}$. The scale-free network is used to describe World Wide Web, scientist's collaboration, actors in a movie etc. An example is illustrated in Figure \ref{FIGURE1}. The network consists of $|n|=490$ nodes. Figure \ref{FIGURE2} shows the occurrence probability distribution $P(d)$ of nodal degree$^{[5]}$ $d$. The horizontal axis is normalized. It is the nodal degree divided by the average degree ($d/\mu(d)$). It is governed by a power low in Equation (\ref{powerlaw}).

\begin{equation}
P(d) \propto d^{-2.1}.
\label{powerlaw}
\end{equation}

The average degree is $\mu(d)=3.6$. The node-to-node deviation in the nodal degree is large. Gini coefficient is also large. It indicates a center-and-periphery structure. Human eyes identify about 10 big hub nodes in the figure easily. The hub nodes influence the way the network operates, communication disseminates, and information stored in the network is accessed. A self-organizing community is often inhomogeneous. A purposefully organized business team is often hierarchical and tends to be inhomogeneous. Relevant empty spots in such organizations are likely to be the hidden nodes. It is relatively easy to obtain a clue on the hidden hub nodes because of their large activeness in communication. Such a problem was studied in $[12]$.

\dsee{3.2~~Homogeneous network}

\begin{figure}
\begin{center}
\includegraphics[scale=0.5,angle=-90]{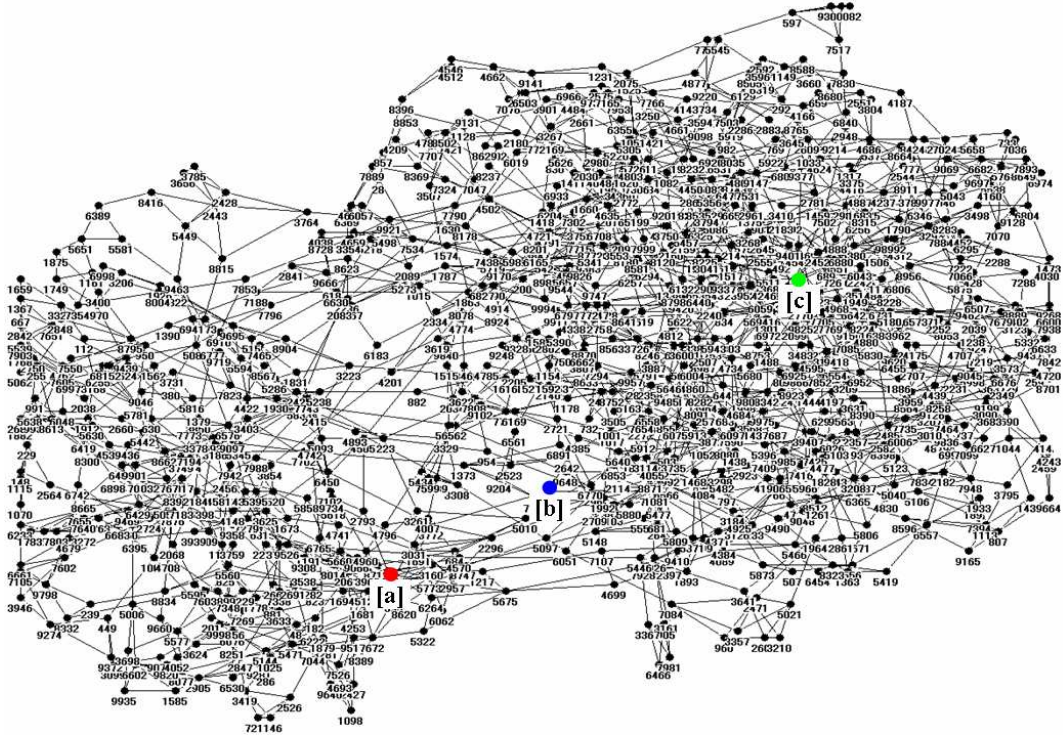}
\end{center}
\caption{Example of a homogeneous social network consisting of $|n|=995$ nodes. The nodes indicated by [a] (red circle), [b] (blue), and [c] (green) are used in the simulation study in 5. The node [a] has the largest nodal degree. The node [b] has the smallest standard deviation of the distance to the other nodes. The node [c] has the smallest mean of the distance.}
\label{FIGURE3}
\end{figure}

\begin{figure}
\begin{center}
\includegraphics[scale=0.3,angle=-90]{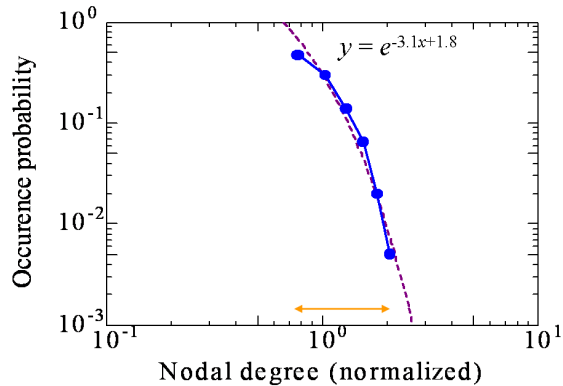}
\end{center}
\caption{Occurrence probability distribution $P(d)$ of the nodal degree $d$ as a function of the normalized degree $d/\mu(d)$.}
\label{FIGURE4}
\end{figure}

A typical homogeneous network is illustrated in Figure \ref{FIGURE3}. The network consists of $|n|=995$ nodes. Figure \ref{FIGURE4} shows the occurrence probability distribution $P(d)$ of nodal degree$^{[5]}$ $d$. It is governed by an exponential law in Equation (\ref{exponentiallaw}).

\begin{equation}
P(d) \propto e^{-3.1d}.
\label{exponentiallaw}
\end{equation}

The degree ranges from 3 to 8. The average degree is $\mu(d)=3.9$. The deviation in the degree is very small. It is the characteristics of the homogeneous network. A small-world network (Watts-Strogatz model) is homogeneous compared with the Barab\'{a}si-Albert model. The Watts-Strogatz model does not possess big hub nodes but short-cut links. It looks difficult to distinguish one node from another from the local topological properties such as the nodal degree.

Such characteristics are suitable for terrorist and criminal organizations$^{[8], [9]}$. Absence of hub nodes is disadvantageous in communication and search efficiency$^{[1]}$, but advantageous in security against exposure and detection$^{[2], [15]}$. It looks a big technical challenge to obtain a clue on the hidden node by identifying relevant empty spots within the communication records observed as a social interaction. A homogeneous network is of particular interest. This network is used in the simulation study in 5.

\dse{4~~Method}

We present our method to discover the empty spots. We employ a heuristic predictor function which evaluates the likeliness of the individual data being an empty spot. Heuristic predictor function is suitable to handle a very large amount of data observed in a very large network. The input to the method is the records $D$, observed for nodes. The records are the collection of the data in the form of baskets in Equations (\ref{BAS1}) and (\ref{BAS2}). The content of the basket $b_{j}$ are a set of nodes $n_{i}$ which are observed simultaneously, or grouped under a specific subject. The number of the baskets is denoted by $|b|$. The number of the nodes in a basket $b_{i}$ is denoted by $|b_{i}|$. The number of species of the nodes is denoted by $|n|$.

\begin{equation}
D=\{b_{j}\} \ (0 \leq j \leq |b|-1).
\label{BAS1}
\end{equation}

\begin{equation}
b_{j} = \{ n_{i} \} \ (0 \leq i \leq |b_{j}|-1).
\label{BAS2}
\end{equation}

The output from the method is a clue on empty spots generated by the predictor function. More specifically, our aim is to identify the basket $b_{i}$ which is related to the empty spots (or the underlying hidden persons) the most likely. The core of our method is, therefore, to design a predictor function $W(b_{i}|D)$ to evaluate the likeliness of the individual baskets $b_{i}$. The basket $b_{i}$ evaluated as the most likely should include the hidden node $n_{x}$, and arise from the links $r_{xj}$ between the node $n_{x}$ and a gateway node $n_{j}$. The gateway node is the observed node which is a neighbor of the hidden node. 

At first, the nodes in the observation are clustered into groups based on the inter-node distance. The distance (or closeness) are defined according to the co-occurrence frequency between the nodes. Occurrence frequency of a node $F(n_{i})$ is defined by Equation (\ref{FRQ}) using a Boolean function $B(s)$ for a proposition $s$ in Equation (\ref{CNT}).

\begin{equation}
F(n_{i}) \equiv \sum_{j=0}^{|b|-1} B(n_{i} \in b_{j}).
\label{FRQ}
\end{equation}

\begin{equation}
B(s) = \left \{ \begin{array}{ll}
                    1 & \mbox{\ \ if $s$ is TRUE} \\
                    0 & \mbox{\ \ otherwise}
                \end{array}
       \right ..
\label{CNT}
\end{equation}

The frequency is the number of the baskets where $n_{i}$ appears. The frequency is increased by 1 when $n_{i}$ appears once or more in a single basket. We use Jaccard's coefficient defined by Equation (\ref{JAC1}) as a measure of the co-occurrence. Jaccard's coefficient is used widely in link discovery, web mining, or text processing$^{[15]}$. Co-occurrence frequency or dependence coefficient may also be used instead of Jaccard's coefficient.

\begin{equation}
J(n_{i},n_{j}) \equiv \frac{F(n_{i} \cap n_{j})}{F(n_{i} \cup n_{j})}.
\label{JAC1}
\end{equation}

Equation (\ref{JAC1}) is converted into Equations (\ref{JAC2}) using Equation (\ref{FRQ}).

\begin{equation}
J(n_{i},n_{j}) = \frac{\sum_{k=0}^{|b|-1} B( (n_{i} \in b_{k}) \wedge (n_{j} \in b_{k})) }{ \sum_{k=0}^{|b|-1} B( (n_{i} \in b_{k}) \vee (n_{j} \in b_{k}) ) }.
\label{JAC2}
\end{equation}

We employ k-medoid clustering algorithm$^{[6]}$ because the amount of necessary calculation is small. It is simple and efficient. It is an EM (expectation-maximization) algorithm similar to k-means algorithm for numerical data. A medoid node $n_{{\rm med}(j)}$ is a node locating most centrally within a cluster $c_{j} \ (0 \leq j \leq |c|-1)$. They are initially selected at random. Other $|n|-|c|$ nodes are classified into the clusters based on the closeness to the medoids. The number of clusters is denoted by $|c|$. Then, a new medoid is selected within the individual cluster so that the sum of closeness from nodes within the cluster to the modoid is maximal. The sum of closeness ($M(c_{j})$ within $c_{j}$) is evaluated by Equation (\ref{Cl1}). Iteration of updating the medoid  $n_{{\rm med}(j)}$ in the cluster to maximize $M(c_{j})$ is carried out until the medoids converge. The resulting clusters are denoted by $c_{j}$. Self-organizing map$^{[6]}$ and other unsupervised learning techniques are the alternatives.

\begin{equation}
M(c_{j}) \equiv \sum_{(n_{i} \in c_{j}) \wedge (n_{i} \neq n_{{\rm med}(j)})} J(n_{{\rm med}(j)},n_{i}).
\label{Cl1}
\end{equation}

Then, the predictor function $W(b_{i}|D)$ in Equation (\ref{W1}) is used to evaluate the likeliness of the individual baskets $b_{i}$ as a candidate which should have included empty spots. The empty spots arise from the hidden participants to the basket, which is the origin of attraction in the empty spots among clusters. The baskets ranked more highly are retrieved by the baskets.

\begin{equation}
W(b_{i}|D) \equiv \frac{1}{|c|} \sum_{j=0}^{|c|-1} \max_{n_{k} \in c_{j}} \frac{B(n_{k} \in b_{i})}{ \sum_{l} B(n_{k} \in b_{l})}.
\label{W1}
\end{equation}

Equation (\ref{W1}) is converted into a simpler formula (Equation (\ref{W2})) using Equation (\ref{FRQ}).

\begin{equation}
W(b_{i}|D) \equiv \frac{1}{|c|} \sum_{j=0}^{|c|-1} \min_{(n_{k} \in c_{j}) \wedge (n_{k} \in b_{i})} F(n_{k}).
\label{W2}
\end{equation}

The predictor function method ranks the baskets by the largeness of the value of $W(b_{i}|D)$. More highly ranked baskets are retrieved when the number of baskets to retrieve is given.

\dse{5~~Evaluation}
\label{evaluation}

We study how precisely the heuristic predictor function method extracts information on the empty spots from the test data set generated as the observed communication records. Communication is a typical social interaction. The homogeneous social network in Figure \ref{FIGURE3} is employed as a model for the communication patterns among 995 persons. We use precision as a measure of the performance. In information retrieval, precision has been used as evaluation criteria, which is the fraction of the amount of relevant data to the amount of the all data returned by search (the data ranked highly by the heuristic predictor function). 

\dsee{5.1~~Test data set}
\label{test}

The test data set is generated in the two steps below. Note that the input communication records $b_{i}$ (representing the observation including the empty spot) are different from the communication patterns $\beta_{i}$ (representing the original communication). The difference, however, does not affect the communication patterns. This difference between $b_{i}$ and $\beta_{i}$ is the target to predict by our method.

\begin{itemize}

\item{} Communication patterns are simulated and formatted into basket-shaped observation records. A basket $\beta_{i}$ includes the nodes which are close to each other. For example, we can imagine a situation where a person starts talking and a conversation takes place among neighboring persons. The area of such influence is specified approximately with the distance from a node. The distance can be measured in the number of hop count. One hop is as long as one link on the network. An example basket is $\beta_{0}$ = \{$n_{954}$, $n_{1930}$, $n_{3261}$, $n_{5093}$, $n_{5223}$, $n_{7743}$, $n_{7808}$, $\dots$\}, representing communication around the node $n_{954}$ in Figure \ref{FIGURE3}.

\item{} Test data set representing observation records (with a hidden node) $b_{i}$ are generated by deleting the nodes of interest $n_{x}$ from the baskets $\beta_{i}$. The deleted nodes and the links connecting them to other nodes become the hidden structure. The baskets $b_{i}$ are the input to the predictor function $W(b_{i}|D)$ in Equation (\ref{W1}). The hidden nodes may exist in multiple baskets. The predictor function ranks and tries to retrieve all of $b_{i}$, which are different from the $\beta_{i}$, as candidates including the hidden nodes and indicating the empty spot. The example basket $\beta_{0}$ results in $b_{0}$ = \{$n_{954}$, $n_{1930}$, $n_{3261}$, $n_{5093}$, $n_{7743}$, $n_{7808}$, $\dots$\} when the node $n_{5223}$ is configured to be the hidden node to predict. The baskets $b_{i}$ are like records of email senders and receivers which lacks in persons expressing their opinions in the oral communication, or like records of persons of on-line chat meetings which lacks in persons using a satellite telephone.

\end{itemize}

In the simulation, we made up 995 baskets ($\beta_{0}$ to $\beta_{994}$) consisting of nodes within a few hops from the initiator node in the 1st step. The number of nodes within 5 hops is about 20\% of the whole nodes on average. This is a relatively long-distance communication.

\begin{figure}
\begin{center}
\includegraphics[scale=0.4,angle=-90]{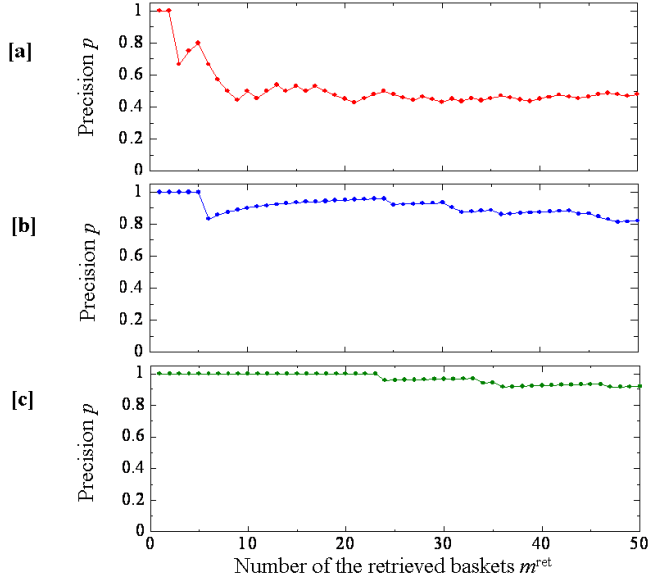}
\end{center}
\caption{Precision $p$ for three trial cases. It is the accuracy to predict the baskets ($b_{i}^{{\rm ret}}$) where nodes were deleted ($b_{i}^{{\rm ret}} \neq \beta_{i}$) as a function of the number of the retrieved baskets ($m^{{\rm ret}}$). The hidden nodes are either nodes around the node [a], [b], or [c] in Figure \ref{FIGURE3}.}
\label{FIGURE5}
\end{figure}

\dsee{5.2~~Precision evaluation}
\label{precision}

We evaluate precision of the output of the method in the three trial cases. In each case, about 10 nodes around the node labeled [a], [b], or [c] in Figure \ref{FIGURE3} are configured as the empty spot. The node [a] has the largest nodal degree. The node [b] has the smallest standard deviation of the distance (hop counts to travel) to the other nodes. The node [c] has the smallest mean of the distance to the other nodes. The nodes [b] and [c] occupy a unique position in terms of a global network topology. They can reach most nodes in the network impartially and efficiently. They are like terrorism sponsors having many (direct or indicrect) relations to the regional activist groups all over the world. Note that the nodal degree is similar in a homogeneous network (actually, 3 to 8 in Figure \ref{FIGURE3}), and that the node [a] is not particularly unique though its degree is the largest.

Precision $p$ is defined as a function of $m^{{\rm ret}}$ by Equation (\ref{PRE}). The number of the baskets retrieved by the heuristic predictor function is denoted by $m^{{\rm ret}}$. It can range from 1 to $|b|$. Individual retrieved baskets are denoted by $b_{i}^{{\rm ret}} \ (0 \leq i \leq m^{{\rm ret}}-1)$. The basket $b_{0}$ is the most likely to include the hidden nodes. The basket $b_{|b|-1}$ is the least likely.

\begin{equation}
p(m^{{\rm ret}}) = \frac{\sum_{i=0}^{m^{{\rm ret}}-1} B(b_{i}^{{\rm ret}} \neq \beta_{i})}{m^{{\rm ret}}}.
\label{PRE}
\end{equation}

Equation (\ref{PRE}) is the ratio of the number of correct baskets to the number of the retrieved baskets. The correct baskets are those where the nodes had been deleted in the 2nd step in 5.1. It means that if $b_{i}^{{\rm ret}} \neq \beta_{i}$, the retrieval is correct. Precision should be 1, and decreases gradually as $m^{{\rm ret}}$ increases if the method works properly. That is, precision is a monotonically decreasing function.

Figure \ref{FIGURE5} shows the evaluated precision $p(m^{{\rm ret}})$ for three trial cases. The top graph is the case when nodes around the node [a] are the hidden nodes. The middle and bottom graphs are for the node [b] and [c]. The horizontal axis is the number of the retrieved baskets ($m^{{\rm ret}}$). The order of the retrieved baskets is according to the largeness of the value which the predictor function outputs. Precision is very good when we try to predict the baskets in case of the nodes [b] and [c]. They occupy a unique position in terms of a global network topology (impartiality and efficiency in communication). The predictor function is suitable to discover the hidden nodes and related links.

On the other hand, precision degrades more steeply as $m^{{\rm ret}}$ increases in case of the node [a]. The results indicate that the method can provide relevant information on the nodes which are hidden but relevant globally in a homogeneous network even if their local characteristics look similar. The nodes included in the retrieved baskets are likely to be the gateway nodes to the hidden nodes. From this information, we may be able to make a detailed investigation plan from the gateway persons toward the hidden persons.

\dse{6~~Concluding Remark}

This contribution presented a heuristic predictor function method which is suitable to predict relevant empty spots in social interaction. The empty spot refers to an empty hard-to-fill space which can be found in the records of the social interaction. It is the clue to the persons in the underlying social network who do not appear in the records. Simulation experiment was demonstrated. This contribution, in particular, focused on the social interaction in homogeneous networks, where we believe that the problem to predict the empty spots is more difficult than that in inhomogeneous networks$^{[12]}$. A test data set in the form of market baskets was generated from the simulated communication patterns. The baskets related to the hidden persons could be predicted by the value of the predictor function accurately.

The idea of relevant empty spots in social interaction can be generalized to hidden but relevant items working in a complex interacting system. We are seeking the possibility where something very new or just emerging is recognized, as well as the possibility where something hidden spatially is discovered. Y. Maeno {\it et al.} studied application to creative thinking$^{[13]}$. An experiment was carried out to see if we can invent a new technical idea from existing technical expertise forest (for example, patent document database or scientific article database). Texts of the patent documents for production machinery were analyzed morphologically and clustered into individual technical expertise. Sensor technology, conveyer system or marking device are example clusters. Participants were engineers and marketing specialists of a manufacturing company. They discussed about the identity of a new technical means (an empty spot) locating near the clusters. A few interesting and practical ideas were invented. Y. Maeno {\it et al.} also studied application to designing catalyst personality fostering mutual understanding and communication among groups indicating opposing preference$^{[11]}$.

We are combining these experiences in many application fields into a new method named human-interactive annealing$^{[10]}$. It is designed to induce discovery from the difference between the individual human's prior understanding of the problem, and the computer's analysis and visualization of the observed data on the problem. Predicting relevant empty spots in social interaction is important in crime investingation, organization design, business engineering, research and development. The method is expected to exploit many unobserved aspects of the social problems.

\rfne

\rf{1}{L. A. Adamic, R. M. Lukose, A. R. Puniyani, and B. Huberman, Search in power-law networks, {\it Physical Review E}, 2001, {\bf 64}: 046135.}

\rf{2}{R. Albert, and H. Jeong, and A. L. Barab\'{a}si, Error and attack tolerance of complex networks, {\it Nature}, 2000, {\bf 406}, 378--381.}

\rf{3}{A. L. Barab\'{a}si, R. Albert, and H. Jeong, Mean-field theory for scale-free random networks, {\it Physica A}, 1999, {\bf 272}: 173--187.}

\rf{4}{P. J. Carrington, J. Scott, S. Wasserman, {\it Models And methods in social network analysis (Structural analysis in the social sciences)}, Cambridge University Press, 2005.}

\rf{5}{L. C. Freeman, Centrality in networks I. Conceptual clarification, {\it Social Networks}, 1979, {\bf 1}: 215--239.}

\rf{6}{T. Hastie, R. Tibshirani, and J. Friedman, {\it The elements of statistical learning: Data mining, inference, and prediction (Springer series in statistics)}, Springer-Verlag, 2001.}

\rf{7}{P. S. Keila, and D. B. Skillicorn, Structure in the Enron email dataset, {\it Journal of Computational \& Mathematical Organization Theory}, 2006, {\bf 11}: 183--199.}

\rf{8}{P. Klerks, The network paradigm applied to criminal organizations, {\it Connections}, 2002, {\bf 24}: 53--65.}

\rf{9}{V. E. Krebs, Mapping networks of terrorist cells, {\it Connections}, 2002, {\bf 24}: 43--52.}

\rf{10}{Y. Maeno, and Y. Ohsawa, Human-computer interactive annealing for discovering invisible dark events, {\it IEEE Transactions on Industrial Electronics}, 2007, {\bf 54}: 1184--1192.}

\rf{11}{Y. Maeno, Y. Ohsawa, and T. Ito, Catalyst personality for fostering communication among groups with opposing preference, {\it Lecture Notes in Artificial Intelligence}, Springer-Verlag, 2007, {\bf 4570}: 806--812.}

\rf{12}{Y. Maeno, and Y. Ohsawa, Stable deterministic crystallization for discovering hidden hubs, {\it Proceedings of the IEEE International Conference on Systems, Man \& Cybernetics, Taipei}, 2006: 1393--1398.}

\rf{13}{Y. Maeno, K. Ito, K. Horie, and Y. Ohsawa, Human-interactive annealing for turning threat to opportunity in technology development, {\it Proceedings of the IEEE International Conference on Data Mining, Workshops, Hong Kong}, 2006: 714--717.}

\rf{14}{Y. Maeno, and Y. Ohsawa, Node discovery problem for a social network, {\it e-print} arxiv.org/abs/0710. 4975, 2007.}

\rf{15}{C. Morselli, C. Giguere, and K. Petit, The efficiency/security trade-off in criminal networks, {\it Social Networks}, 2007, {\bf 29}, 143--153.}

\rf{16}{Y. Ohsawa, Data crystallization: chance discovery extended for dealing with unobservable events, {\it New Mathematics and Natural Computation}, 2005, {\bf 1}: 373--392.}

\rf{17}{{\it Emergent information technologies and enabling policies for counter-terrorism} (ed. by R. L. Popp and J. Yen), IEEE Press, 2006.}

\rf{18}{S. Singh, J. Allanach, T. Haiying, K. Pattipati, and P. Willett, Stochastic modeling of a terrorist event via the ASAM system, {\it Proceedings of the IEEE International Conference on Systems, Man \& Cybernetics, Hague}, 2004, {\bf 6}: 5673--5678.}

\rf{19}{D. J. Watts, and S. H. Strogatz, Collective dynamics of small-world networks, {\it Nature}, 1998, {\bf 398}: 440--442.}

\end{document}